\def\year{2019}\relax
\documentclass[letterpaper]{article} 
\usepackage{aaai19}  
\usepackage{times}  
\usepackage{helvet}  
\usepackage{courier}  
\usepackage{url}  
\usepackage{graphicx}  
\usepackage{eurosym}

\usepackage{caption}	
\usepackage{subcaption}
\newcommand{\citeaby}[1]{\citeauthor{#1} \shortcite{#1}}	
\title{Strategic Features for General Games}

\author{Cameron Browne, Dennis J. N. J. Soemers, and Eric Piette \\ 
Department of Data Science and Knowledge Engineering (DKE) \\ 
Maastricht University, Bouillonstraat 8-10 \\
Maastricht, 6211 LH, The Netherlands
}

\begin{document}

\maketitle


\begin{abstract}
This short paper describes an ongoing research project that requires the automated self-play learning and evaluation of a large number of board games in digital form. 
We describe the approach we are taking to determine relevant features, for biasing MCTS playouts for arbitrary games played on arbitrary geometries. 
Benefits of our approach include efficient implementation, the potential to transfer learnt knowledge to new contexts, and the potential to explain strategic knowledge embedded in features in human-comprehensible terms.
\end{abstract}


\section{Introduction}

The Digital Ludeme Project\footnote{Funded by a \euro2m ERC Consolidator Grant (http://ludeme.eu).} is a five-year research project, recently launched at Maastricht University, that aims to model the world's traditional strategy games in a single, playable digital database. 
This database will be used to find relationships between games and their component {\it ludemes},\footnote{Units of game-related information~\cite{Borvo1975}.} in order to develop a model of the evolution of games throughout recorded human history and to chart their spread across cultures worldwide. 
This paper describes a new approach to defining strategic features for general games that we are implementing in order to achieve the project's aims.


\subsection{Statement of the Problem}

While there exists substantial archaeological evidence for {\it ancient games} (i.e. those played before 500{\sc ad}) the rule sets explaining how these games were actually played are typically lost; our modern understanding of ancient and early games is based primarily on modern reconstructions that have proved to be less than reliable~\cite{Murray1951}. 

A key challenge in this project will be to evaluate reconstructions of ancient rule sets and to improve them where possible. 
The general game system we are developing for this project, called {\sc Ludii}, must therefore be able to:

\begin{enumerate}

\vspace{-1mm}
\item Model the full range of traditional strategy games.

\vspace{-1mm}
\item Play them at a competent human level.

\vspace{-1mm}
\item Evaluate rule set reconstructions for: a) historical authenticity, and b) quality of play.

\vspace{-1mm}
\item Improve rule set reconstructions where needed.

\end{enumerate}

We aim to model the 1,000 most influential traditional games throughout history, each of which may have multiple interpretations, each of which may require hundreds of variant rule sets to be tested. 
This is therefore not just a mathematical/computational challenge but also a logistical one, as {\sc Ludii} may need to be able to learn to play and evaluate up to a million candidate rule sets over the project's five years. 

This paper outlines work-in-progress for achieving this ambitious goal, through the self-play learning of simple lightweight features that embody strategic knowledge about the games being modelled. 
The features we propose are geometry-independent, facilitating their transfer to new contexts, and represent basic strategies that may be understood and explained in human-comprehensible terms.


\section{Geometric Piece Features}  

The {\sc Ludii} general game system employs {\it Monte Carlo tree search} (MCTS)~\cite{Browne2012} as its core method for AI move planning, which has proven to be a superior approach for general games in the absence of domain-specific knowledge~\cite{Finnsson2010}. 
While MCTS performance can vary significantly if strictly random playouts are used~\cite{Browne2013}, the reliability of Monte Carlo simulations can be improved by biasing playouts with domain-relevant features. 

There is a significant amount of prior research that describes the use of local features or patterns for game-playing agents tailored towards a specific game. 
Some of the most common approaches for generating features are:

\begin{itemize}

\item Manually constructing features using expert knowledge. This was, for example, done by \citeaby{Gelly2006ModificationUCT} for the game of Go, and by \citeaby{Sturtevant2007FeatureConstructionHearts} for the game of Hearts.

\item Exhaustively generating all possible patterns up to a certain restriction (often limited to, e.g., areas of $3$$\times$$3$ cells). This was, for example, done by~\citeaby{Gelly2007CombiningOnlineOffline} for the game of Go, and~\citeaby{Lorentz2017PatternsBreakthrough} for the game of Breakthrough. \citeaby{Sturtevant2007FeatureConstructionHearts} also exhaustively generated combinations (for instance pairs, triples, etc.) of the manually selected features for the game of Hearts as described above.

\item Supervised learning based on databases of human expert games. 
This has been particularly common in the game of Go~\cite{Enderton1991GolemGoProgram,Stoutamire1991MachineLearningGamePlayGo,vanderWerf2003LocalMovePredictionGo,Bouzy2005BayesianGeneration,Stern2006BayesianPatternRanking,Araki2007MovePredictionGo,Coulom2007EloRatingsPatterns}.

\end{itemize}

Other approaches include Gradual Focus \cite{Skowronski2009AutomatedDiscovery}, a genetic programming approach to find useful patterns \cite{Hoock2010BanditBasedGenProg}, and the use of deep convolutional neural networks for learning relevant features through self-play~\cite{SilverHuangEtAl16nature}.  

\begin{figure}[htbp]
\centering
\includegraphics[width=.7\linewidth]{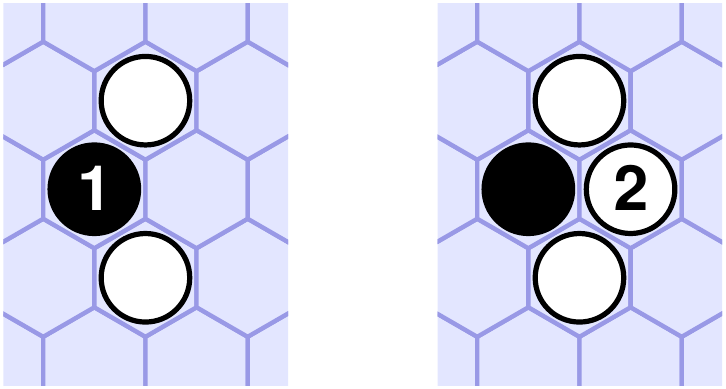}
\caption{Bridge completion is a beneficial pattern in Hex.}
\label{fig:HexBridge}
\end{figure}

The features we are interested in here are geometric piece patterns of arbitrary topology. 
For example, Figure~\ref{fig:HexBridge} shows a pattern of pieces that represents a beneficial move for connection games played on the hexagonal grid. 
If Black has just played move {\bf 1} to intrude into the virtual connection of two White pieces (left), then White should reply immediately with move {\bf 2} to complete that connection. 
Biasing MCTS playouts to play such moves with higher probability was found to dramatically improve AI playing strength for connection games such as Hex and Y~\cite{Raiko}.


\section{Feature Definition}  \label{sec:FeatureDefinition}

Each feature $x$ defines:

\begin{enumerate}

\item A \textit{pattern} $p_x$ which specifies one or more elements that must be present or absent in certain locations in game states for feature $x$ to hold.

\item An action $a_x$ which may be encouraged or discouraged in game states where $x$ holds.

\end{enumerate}

For most board games, actions can be sufficiently specified using one or two integers; a location to move ``to'', and in some games a location to move ``from''. 
For example, in games such as Hex and Go a single board location is sufficient to uniquely identify any action in any particular game state. 
In games such as Draughts or Chess, the current location of the piece to be moved is also required.


\subsection{Syntax}

Each pattern consists of a set of {\it elements}.  
We define the following element types for each potential feature element, relative to the current mover:

\begin{itemize}

\vspace{-1mm}
\item ``-'' Off-board location (for locating edges and corners).

\vspace{-1mm}
\item ``.'' Empty board location.

\vspace{-1mm}
\item ``o'' Friendly piece (i.e. of the mover's colour).

\vspace{-1mm}
\item ``x'' Enemy piece (i.e. not of the mover's colour).

\vspace{-1mm}
\item ``P{\it n}'' Piece belonging to player {\it n}.

\vspace{-1mm}
\item ``I{\it n}'' Piece with index {\it n} in the game definition.\footnote{The game definition maintains a unique index for each type of equipment it involves, including {\it containers} (boards, piles, hands, decks, etc.) and {\it components} (pieces, tiles, dice, cards, etc).}

\end{itemize}

We also define ``!'' (i.e. ``not'') versions of these element types. 
For example, ``!P2'' means that the specified location does not contain a piece belonging to Player 2. 
Locations can simultaneously have multiple qualifiers, e.g. a location with both ``x'' and ``!I3'' qualified would match an enemy piece at that location unless it has index 3 in the game definition. 


\subsection{Topology}

Rather than tying feature topology to any particular board or grid type, feature elements are described by their {\it relative locations} on the underlying graph of the game board (or more accurately its dual, which defines orthogonal cell adjacencies). 
Starting with a given {\it anchor} location and direction, the relative location of each element is defined as a {\it walk} through the underlying board graph. 
We assume that the orthogonal neighbours of each board location are listed in consecutive clockwise order, including null placeholders for off-board steps.

Each walk is defined as a sequence of adjacent {\it steps} through the underlying board graph, where:

\begin{itemize}

\vspace{-1mm}
\item 0 denotes a step forwards in the current direction.

\vspace{-1mm}
\item $-\frac{1}{3}$ denotes $\frac{a}{3}$ counterclokwise turns in a cell with $a$ sides (i.e. one turn in a triangle), then a step forwards, 

\vspace{-1mm}
\item $+\frac{2}{4}$ denotes $\frac{2a}{4}$ clockwise turns in a cell with $a$ sides (i.e. two turns in a square), then a step forwards, etc.

\end{itemize} 

\begin{figure}[htbp]
\centering
\includegraphics[width=1\linewidth]{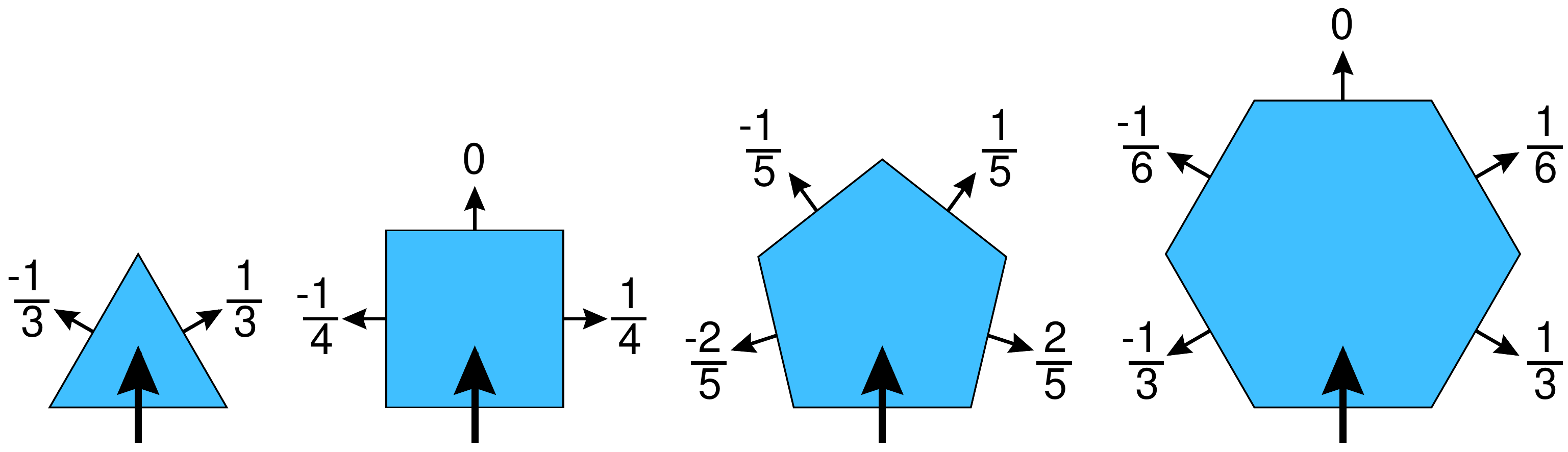}
\caption{Steps through various cells based on CW turns.}
\label{fig:Steps}
\end{figure}

Figure~\ref{fig:Steps} shows the relative steps through cells with different numbers of sides. Fractional turns are rounded to the nearest sensible fraction for any given cell (e.g., a turn of $\frac{1}{4}$ in a triangle becomes equal to a turn of $\frac{1}{3}$). The basic mechanism is similar in principle to the use of {\it turtle commands} in computer graphics. 

Note that cells with an odd number of sides, such as the triangular and pentagonal cells shown in Figure~\ref{fig:Steps}, do not have a forwards (0) direction. 
Such cases may be handled by instantiating two pattern instances per ambiguous step: one with $0$ replaced by $-\frac{1}{a}$ and one with $0$ replaced by $+\frac{1}{a}$ for that step in a cell with $a$ sides.

\begin{figure}
\centering
\includegraphics[width=.9\linewidth]{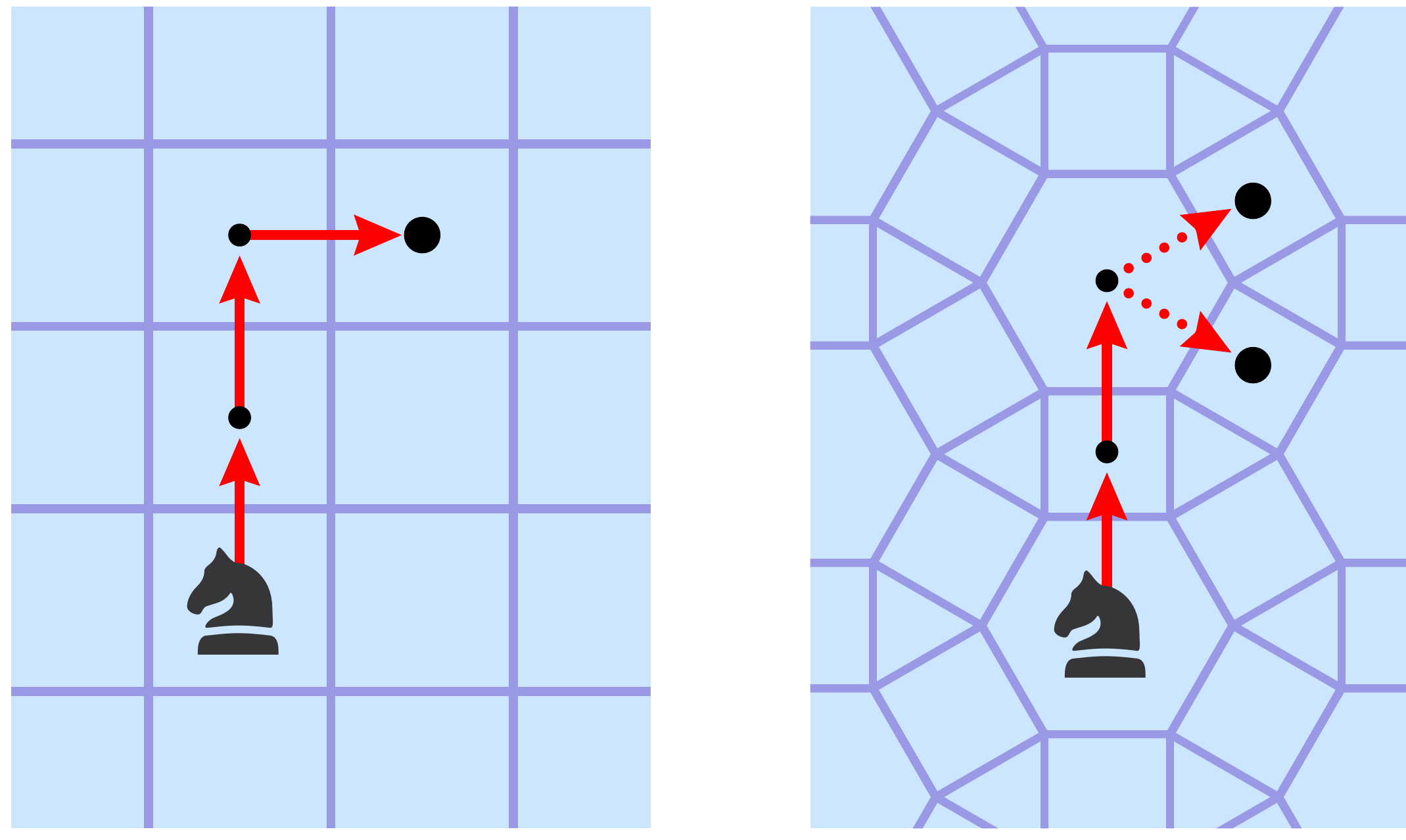}
\caption{Relative knight move $\{0,0,\frac{1}{4}\}$ on the square grid (left) and two equivalent moves on the semi-regular 3.4.6.4 grid (right).}\label{fig:KnightSteps}
\end{figure}

For example, a knight move may be described using this notation as $\{0,0,\frac{1}{4}\}$, as shown in Figure~\ref{fig:KnightSteps} (left). 
This description can be mapped directly to other grids of arbitrary topology to produce plausible results, such as the semi-regular 3.4.6.4 tiling (right).
Note that the turn of $\frac{1}{4}$ is ambiguous in the semi-regular grid, and is therefore split into two possible patterns.

\subsection{Local vs Global Patterns}

Patterns may be defined as either:

\begin{itemize}

\item {\it Relative:} Apply to all valid anchor locations across the board (taking null locations into account).

\item {\it Absolute:} Apply only to the specified anchor location.

\end{itemize}

For convenience, the number of valid rotations and reflections for each pattern may also be specified, so that all possible instances of a given pattern may be generated across the board from a single description.  


\subsection{Reactive vs Proactive Patterns}

We distinguish between {\it reactive} patterns that trigger actions directly in response to the previous player's last move,\footnote{There may be more than one opponent.} and {\it proactive} patterns that can trigger actions anywhere across the board regardless of the previous player's last move. 
This distinction is similar to the distinction between ``responsive'' and ``non-responsive'' patterns used in~\cite{SilverHuangEtAl16nature}. 

\begin{figure}[htbp]
\centering
\includegraphics[width=.9\linewidth]{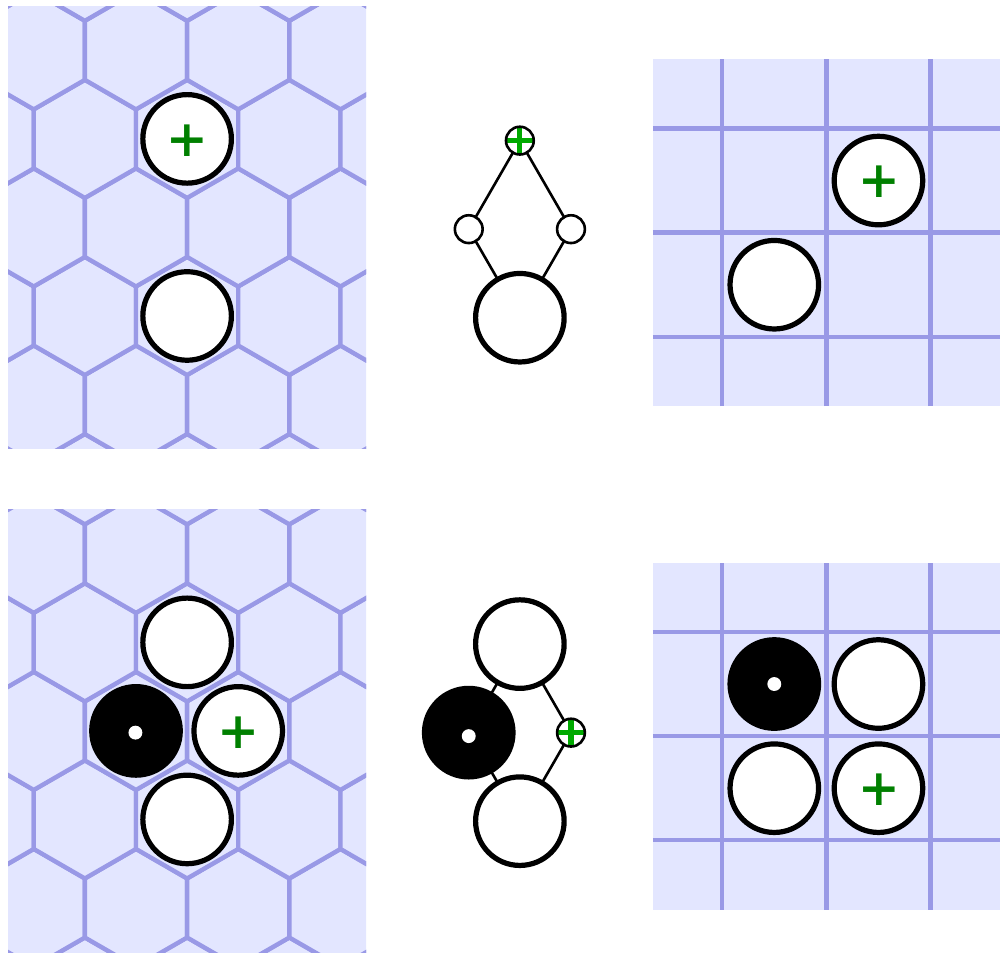}
\caption{Reactive (bottom) and proactive (top) features.}
\label{fig:ReactiveProactive}
\end{figure}

For example, the bottom row of Figure~\ref{fig:ReactiveProactive} shows the Hex pattern described earlier in Figure~\ref{fig:HexBridge} (left) and a reactive pattern that describes this situation (middle). 
The top row shows a proactive pattern that encourages Hex players to play two disjoint steps away from existing friendly pieces. 
For clarity in the diagrams, friendly pieces are denoted as white disks and enemy pieces as black discs (the last move made is dotted). 
White dots indicate empty cells, edges between elements indicate cell adjacency, and green ``+'' symbols indicate the preferred action triggered by each pattern. 
The rightmost column of Figure~\ref{fig:ReactiveProactive} shows how these patterns map easily from the hexagonal grid to the square grid. 

\section{Implementation}

The {\sc Ludii} system and associated feature mechanisms are implemented in Java 8. 
Internal game states are defined using a custom {\tt BitSet} class, called a {\tt ChunkSet}, that compresses the required state information in the minimum number of bits, based on each game's definition. 

For example, if a game involves no more equipment than a board and uniform pieces in $N$ colours, then the game state is described by a {\tt ChunkSet} subdivided into {\it chunks} of $B$ bits per board cell, where $B$ is the lowest power of 2 that provides enough bits to represent every possible state per cell (including state 0 for the empty cells).\footnote{Chunk sizes are set to the lowest power of 2 to avoid issues with chunks straddling consecutive {\tt long} member variables.}  

For each game, we seek to derive an optimal feature set $F$, which is the set of features that outperform all other known feature sets for that game on average, within a given computational budget. 


\subsection{Feature Instantiation}

As each game is loaded, an {\it instance} of every possible valid rotation, reflection and translation is pre-generated once for all component features $f_x$ of that game's optimal feature set $F$, including combinations of valid element types within each feature. 
Each feature definition can therefore generate hundreds of instances. 

Each instance is defined using the same custom {\tt BitSet} class as the game state. 
Hence, each feature instance can be matched to the given game state efficiently using bitwise parallel operations, such that only a few bitwise operations need to be applied per instance match test, regardless of the feature's complexity.


\subsection{Feature Application}

Feature sets are applied to bias MCTS playouts as follows:

\begin{enumerate}

\item For each player move, all legal moves are generated and initialised with equal probability of being selected. 

\item Each reactive feature instance corresponding to the previous player's last move is then checked for a match to the current game state, and if a match is detected then the feature's weight is added to the probability of the feature's action being selected (note that feature weights can be negative for detrimental moves).

\item Each proactive feature instance is then applied across the board and checked for a match to the current game state, and if a match is detected then the feature's weight is added to the probability of the feature's action being selected. 

\item The move to be made is then selected randomly according to the resulting biased distribution.

\end{enumerate}

Features can similarly be used in other parts of MCTS, such as its selection phase. Note that reactive features are much more efficient to apply than proactive ones, as only those patterns relevant to the previous player's last move need be tested. 
We therefore seek to generate reactive features, and to replace proactive features with their reactive equivalents, where possible.


\subsection{Feature Generation}  \label{sec:FeatureGeneration}

Manually generating features using expert knowledge, or extracting them through supervised learning from human expert games, are not suitable solutions for the current task in which a wide variety of unfamiliar games must be supported. 
Exhaustive generation of features can be done more easily, but has clear drawbacks due to the massive volume of features it can generate.

One advantage of describing games by their component ludemes is that these may be exploited to generate a plausible set of candidate feature patterns based on the specified rules and equipment. 
Such patterns may not be suitable features in and of themselves, but represent ``minimum'' patterns which must be present in every feature -- for example, every Hex or Go feature must contain an empty ``to'' action location -- allowing significant reductions in the number of possible patterns to be generated. For many games, this kind of knowledge can be extracted automatically from ludemes.

The subsequent fine-tuning of features to obtain the optimal feature set $F$ for a given game is still a work-in-progress. 
Suffice it to say that {\it frequent pattern mining}~\cite{Aggarwal2014} approaches to iteratively building features from large numbers of randomly generated self-play games has not proved effective. 
Learning feature sets against a random opponent can have disastrous -- and sometimes amusing -- results in the unexpected behaviours that can thwart the random player.  
There does not seem to be any escape from evaluating feature sets using more time-consuming but realistic tests against intelligent AI opponents, with MCTS as the baseline yardstick. 


\section{Features and Strategies}

\begin{figure}[htbp]
\centering
\includegraphics[width=.80\linewidth]{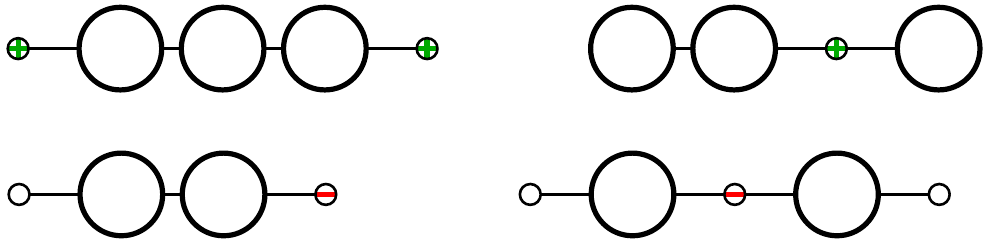}
\caption{Patterns for a ``line of 4'' strategy (top) and a ``not line of 3'' strategy (bottom).}
\label{fig:PatternsLines}
\end{figure}

An attractive aspect of the proposed approach is that learnt features have the potential to encode simple strategies for the games to which they apply. 
For example, the features shown on the top row of Figure~\ref{fig:PatternsLines} encourage the player to make lines of four pieces of their colour, while the features shown on the bottom row encourage the player to {\it not} make lines of three pieces of their colour (the red ``--'' symbols indicate negative weights that discourage such actions). 


\begin{figure}[htbp]
\centering
\includegraphics[width=.62\linewidth]{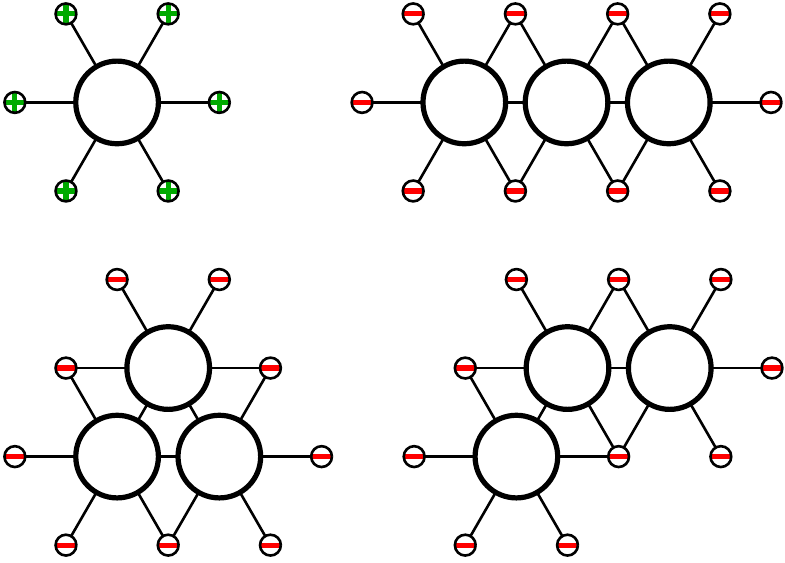}
\caption{Patterns for a ``form groups of 3'' strategy.}
\label{fig:PatternsGroup3}
\end{figure}

\begin{figure}[htbp]
\centering
\includegraphics[width=.545\linewidth]{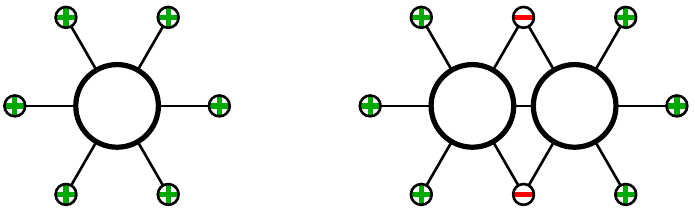}
\caption{Patterns for a ``form long thin groups'' strategy.}
\label{fig:PatternsThinGroups}
\end{figure}

The features shown in Figure~\ref{fig:PatternsGroup3} encourage the player to form groups of three pieces of their colour, by encouraging singleton pieces to grow in any direction then discouraging growth beyond group size 3. 
Similarly, the features shown in Figure~\ref{fig:PatternsThinGroups} encourage the player to form long, thin groups of their pieces, by encouraging singleton pieces to grow in any direction, then encouraging friendly pairs to extend at the ends but discouraging growth at the common points adjacent to both pieces (which would create shorter, thicker groups). 


\subsection {Feature Explanation}

An additional benefit of the ludemic model for games is that features applicable to a given game may be readily transferred as candidate features for related games defined by similar ludemes. 
Any feature that has proved relevant for a given context is a good starting point for similar contexts. 

Further, we can potentially invert the causal relationship between ludemes and features to reverse engineer comprehensible explanations for the learnt strategies that optimal feature sets represent, by determining {\it which} ludemes are responsible for given features. 
This relationship between a game's ludemes and its derived features is further strengthened by the fact that initial candidate feature sets are derived directly from the game's ludemic description during feature generation process.

Each ludeme in the {\sc Ludii} Ludeme Library corresponds to a Java class with an appropriate name, providing a source of convenient plain English captions for learnt concepts. 
And even if the ludeme names do not provide sufficient explanation in themselves, they provide hints for further post-processing steps to find geometric relationships within the feature patterns. 
It is plausible that simple descriptions such as ``make lines of 4'' or ``make groups of 3'' may be derived from such feature sets. 
Implementing such mechanisms for explainable AI in the context of startegy learning for general games will be a key focus of future work. 


\section{Conclusion}

Lightweight features based on geometric piece patterns have a number of advantages for our work on the Anonymous Project. 
They improve MCTS playing strength, map readily to other geometries, encode simple strategies, can be associated with the underlying ludemic descriptions of games, and have the potential to help explain learnt strategies in human-comprehensible terms. 
We will continue developing and testing this approach as the LUDII general game system matures to support an increasing range of game types, and investigating the automated extraction and explanation of relevant strategies from such learnt features. 


\section{Acknowledgements}

This research is part of the European Research Council-funded Digital Ludeme Project (ERC Consolidator Grant \#771292) being run by Cameron Browne. 
We would also like to acknowledge the RIKEN Institute's Advanced Intelligence Project (AIP), especially Kazuki Yoshizoe, for their generous support of prior research that led to this work. 


\bibliography{References}
\bibliographystyle{aaai}

\end{document}